\documentclass{article}

\usepackage{microtype}
\usepackage{graphicx}
\usepackage{subfigure}
\usepackage{booktabs} 

\usepackage{hyperref}

\usepackage[accepted]{icml2024}

\usepackage{inconsolata}
\usepackage[T1]{fontenc}

\usepackage{amsmath}
\usepackage{amssymb}
\usepackage{mathtools}
\usepackage{amsthm}
\usepackage{xcolor}

\definecolor{olivegreen}{HTML}{006400}
\definecolor{echoblue}{HTML}{0099CC}
\definecolor{gold}{HTML}{d18600}
\definecolor{vividred}{HTML}{E60B42}
\definecolor{echonavy}{HTML}{0054B2}
\definecolor{darkgry}{HTML}{333333}
\definecolor{echopurple}{HTML}{9400D1}

\usepackage[capitalize,noabbrev]{cleveref}

\theoremstyle{plain}

\theoremstyle{definition}

\theoremstyle{remark}

\usepackage[textsize=tiny]{todonotes}

\icmltitlerunning{The CLRS-Text Algorithmic Reasoning Language Benchmark}

\begin{document}

\twocolumn[
\icmltitle{The CLRS-Text Algorithmic Reasoning Language Benchmark}



\icmlsetsymbol{equal}{*}
\icmlsetsymbol{adv}{$\dagger$}

\begin{icmlauthorlist}
\icmlauthor{Larisa Markeeva}{equal,gdm}
\icmlauthor{Sean McLeish}{equal,mary}
\icmlauthor{Borja Ibarz}{equal,gdm}
\icmlauthor{Wilfried Bounsi}{gdm}
\icmlauthor{Olga Kozlova}{gdm}
\icmlauthor{Alex Vitvitskyi}{gdm}
\icmlauthor{Charles Blundell}{gdm}
\icmlauthor{Tom Goldstein}{adv,mary}
\icmlauthor{Avi Schwarzschild}{adv,cmu}
\icmlauthor{Petar Veli\v{c}kovi\'{c}}{adv,gdm}
\end{icmlauthorlist}

\icmlaffiliation{gdm}{Google DeepMind}
\icmlaffiliation{mary}{University of Maryland}
\icmlaffiliation{cmu}{Carnegie Mellon University}

\icmlcorrespondingauthor{Larisa Markeeva}{lmarkeeva@google.com}
\icmlcorrespondingauthor{Sean McLeish}{smcleish@umd.edu}
\icmlcorrespondingauthor{Petar Veli\v{c}kovi\'{c}}{petarv@google.com}

\icmlkeywords{Neural Algorithmic Reasoning, Large Language Models, CLRS, Graph Neural Networks}

\vskip 0.3in
]



\printAffiliationsAndNotice{\icmlEqualContribution} 

\begin{abstract}
Eliciting reasoning capabilities from language models (LMs) is a critical direction on the path towards building intelligent systems. Most recent studies dedicated to reasoning focus on out-of-distribution performance on procedurally-generated synthetic benchmarks, bespoke-built to evaluate specific skills only. This trend makes results hard to transfer across publications, slowing down progress. Three years ago, a similar issue was identified and rectified in the field of neural algorithmic reasoning, with the advent of the \emph{CLRS} benchmark. CLRS is a \emph{dataset generator} comprising graph execution traces of classical algorithms from the Introduction to Algorithms textbook. Inspired by this, we propose \textbf{CLRS-Text}---a textual version of these algorithmic traces. Out of the box, CLRS-Text is capable of procedurally generating trace data for thirty diverse, challenging algorithmic tasks across any desirable input distribution, while offering a standard pipeline in which any additional algorithmic tasks may be created in the benchmark. We fine-tune and evaluate various LMs as generalist executors on this benchmark, validating prior work and revealing a novel, interesting challenge for the LM reasoning community. Our code is available at \url{https://github.com/google-deepmind/clrs/tree/master/clrs/_src/clrs_text}.
\end{abstract}

\section{Introduction}

In spite of the impressive performance of language models (LMs) in a variety of scenarios \citep{reid2024gemini,achiam2023gpt}, they also exhibit some well-documented failures, especially when it comes to \emph{reasoning} problems \citep{dziri2024faith}. Some examples of this relate to reverse concepts \citep{berglund2023reversal}, elementary arithmetic \citep{shen2023positional,zhou2024transformers} and geometry \citep{mouselinos2024beyond}, or recognising higher-order languages \citep{deletang2022neural}. Improving base LM performance on reasoning tasks is important, as their present unreliability makes them unwieldy to use in environments requiring robust behaviours, e.g. multi-step planning and scientific problems. 

Reasoning capabilities of LMs are often evaluated using static datasets requiring mathematical reasoning \citep{cobbe2021training,hendrycks2021measuring} or QA \citep{liu2020logiqa,rein2023gpqa}, but it is well-understood that hill-climbing on any \emph{static} dataset may result in an illusion of progress, especially given vast quantities of data available in pre-training corpora. Even resampling of values on these benchmarks can result in drastic regressions---a phenomenon known as the \emph{reasoning gap} \citep{srivastava2024functional}.

To simulate how a model might adapt to unfamiliar situations, many studies dedicated to improving reasoning in LMs will evaluate out-of-distribution performance (typically \emph{length generalisation}) on synthetic tasks \citep{anil2022exploring,zhou2022teaching,zhou2023algorithms,ruoss2023randomized,zhou2024transformers,sanford2024understanding}. Critically, each of these works constructs a bespoke dataset, making it hard to evaluate the relative importance of various works. We notice that, only three years ago, a very similar conundrum affected the evaluation of algorithmic execution capabilities of graph neural networks \citep{xu2019can,velivckovic2019neural,tang2020towards}. Therein, the issue was successfully addressed with the advent of the \emph{CLRS} benchmark \citep{velivckovic2022clrs}---a dataset generator comprising execution traces of thirty classical algorithms from the Introduction to Algorithms textbook \citep{cormen2022introduction}, while allowing for a principled way to generate traces for novel tasks, enabling several novel research directions \citep{ibarz2022generalist,bevilacqua2023neural,minder2023salsa,jurss2024recursive}.

Our aim is to \emph{bring the benefits of CLRS into language modelling}, yielding the \textbf{CLRS-Text} benchmark. CLRS-Text is a procedural dataset generator based on converting the graph-based traces within CLRS into textual form, making them suitable for ingestion by language models---see Figure \ref{fig:arr}, with additional examples given in Appendix \ref{app:samples}.

\begin{figure*}
    \includegraphics[width=\linewidth]{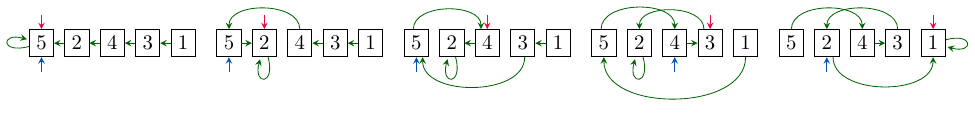}
    \textbf{\texttt{\textcolor{olivegreen}{insertion\_sort:\\
    key: [5 2 4 3 1], initial\_trace: [5 2 4 3 1]\\
    trace | pred:}\\
    \textcolor{echonavy}{{[}2 5 4 3 1{]}, {[}2 4 5 3 1{]}, {[}2 3 4 5 1{]} |} \textcolor{vividred}{{[}1 2 3 4 5{]}}}}
    \caption{{\bf Top:} The \emph{graph} algorithmic trace of insertion sorting a list $[5,2,4,3,1]$ in graph form (reprinted from \citet{velivckovic2022clrs}). {\bf Bottom:} The same algorithmic trace, represented \emph{textually}, by using our provided CLRS-Text generator. The model receives as \emph{input} (depicted in \textcolor{olivegreen}{\bf green}) the input array (\texttt{key}) and the initial value of the sorting trace (\texttt{initial\_trace}), using which it is prompted to predict the \emph{trace} (depicted in \textcolor{echonavy}{\bf blue}) of gradually sorting the list, by inserting one element at a time into a partially sorted list, from left to right. At the end, the model needs to \emph{output} the final sorted array (depicted in \textcolor{vividred}{\bf red}), and it is evaluated on whether this array is predicted correctly.}
    \label{fig:arr}
\end{figure*}

\section{Motivation}

To clarify why we find CLRS-Text to be an important benchmark for evaluating reasoning capabilities, we will first discuss our operational definition of reasoning, along with the implications for what a dataset for evaluating this kind of reasoning might look like.

To us, \emph{reasoning is a \textbf{robust} procedure for solving instances of a problem}. We impose no requirement for this procedure to be fully accurate: we consider humans capable of reasoning, yet human reasoning is often approximate or relies on partial information. Further, we do not require the model to explicitly trace its computation using symbols---reasoning may be successfully done in the latent space. The key factor, \emph{robustness}, implies the model should behave \emph{consistently} across diverse problem instances\footnote{Or, at least, the model should be able to give appropriate confidence estimates in its answers---or even avoid answering altogether---if the input substantially escapes the support of its training data. This is hard to reconcile with modern practices of instruction tuning \citep{wei2021finetuned}, which effectively teaches a model to always attempt to provide answers.}. Accordingly, we care about \emph{out-of-distribution} generalisation in LMs. And as is already known, even for relatively simple arithmetic operations such as multiplication, many present frontier models fail even at modest length generalisation \citep{shen2023positional}.

How can we evaluate out-of-distribution generalisation? Firstly, note that full OOD is hard to encounter or even construct with modern pre-training setups, wherein LMs are trained on Internet-scale data (though we will provide results of pre-trained frontier LMs on CLRS-Text as an indication of their immediate algorithmic reasoning capabilities). Because of this, we argue for hand-crafting specialised data that we will finetune models on, and also construct appropriate IID and OOD evaluation datasets for. This is standard practice in several relevant prior works \citep{deletang2022neural,ruoss2023randomized,shen2023positional}.

In order to be able to do this automatically, we need to focus our attention on tasks that allow outputs to be generated (a) reliably, (b) efficiently, and (c) for any valid input distribution of interest. These constraints, taken together with the expectation that reasoning should be a robust procedure, imply that we should train and evaluate our models on traces of \textbf{polynomial-time algorithms}, which exactly correspond to robust, well-defined procedures that perform their computations in a tractable manner.

This is exactly what the CLRS benchmark is designed to do---expose traces of polynomial-time algorithms in a rigid, efficient and unified manner, for any valid input distribution. And it is this observation that fuels our decision to extend CLRS into the textual domain and train language models on its traces. It is our hope that CLRS-text has the potential to become an important benchmark for reasoning, precisely because it allows for easy generation of bespoke trace data at various distributions, and simplifies setting up comparisons across multiple papers that use it. 

Lastly, and conveniently: since we evaluate our model on procedurally-generated test data, we can constantly \emph{resample} the test datapoints, ameliorating risks of reasoning gaps from hill-climbing static datasets \citep{srivastava2024functional}.

\section{CLRS-Text construction}

As already mentioned, CLRS-Text is entirely based on the representations computed by the CLRS benchmark \citep{velivckovic2022clrs}. For a brief recap: each algorithm in CLRS is specified by \emph{inputs}, \emph{traces}\footnote{In the original CLRS benchmark, the term ``hints'' is occasionally used instead of ``traces''.} and \emph{outputs}, and by default, CLRS offers access to thirty classical algorithms from \citet{cormen2022introduction}, spanning sorting, searching, divide \& conquer, greedy algorithms, dynamic programming, graph algorithms, string algorithms and geometric algorithms.

For each sample of every algorithm, it is assumed that inputs and outputs are kept fixed, whereas the traces represent the trajectory of (internal) states the algorithm goes through while computing the output. Since CLRS was originally designed to train non-autoregressive models, it natively leverages a \emph{graph} representation of this data.

In contrast, CLRS-Text converts the trace data to text. How should this conversion be done? In general, this is something that users of CLRS-Text can tweak to their needs, by modifying the default converters provided in \texttt{clrs\_utils.py}.

The default conversion function we provide (with outputs illustrated by Figure \ref{fig:arr} and Appendix \ref{app:samples}) is designed with \emph{limited context windows} in mind. We would like our traces to completely fit in context of current small-tier models such as Gemma \citep{team2024gemma}, to allow for simplified training at a wide range of parameter scales. This means that, especially for tasks including information on edges of graphs ($O(n^2)$ entries for problems of size $n$), we cannot afford to print all parts of the algorithm's trace, and instead we focus on printing \emph{exactly one variable}'s trace---the variable which eventually converges to the output. In the concrete case of  sorting algorithms, the trace we print is the state of the input array after each step of the algorithm. 

The only algorithm in the thirty default algorithms of CLRS for which we do not provide a trace is the segment intersection algorithm, as it has $O(1)$ time complexity and therefore does not have atomic steps that converge to the final output.

Note that, because we do not print the entirety of the algorithm's state at every step, it is possible that the trace may remain \emph{unchanged} in certain steps---for example, when insertion sorting an already-sorted array of $n$ elements, each of the $n$ steps of the trace will be identical. We argue that it is \emph{useful} to encourage the model to produce such traces, as it explicitly indicates to the model the likely ``thinking time'' needed for solving the task through chain of thought \citep{wei2022chain}. Further, recent results indicate that the mere amount of chain of thought tokens may be correlated with the relative increase in expressive power of Transformer-based language models \citep{merrill2023expresssive}.

Lastly, while CLRS-Text provides the same thirty algorithms present in CLRS by default, because of the strong synergy of the two benchmarks, adding new algorithmic tasks to CLRS-Text's generator requires no significant added effort compared to adding them to CLRS.

The reader interested in adding new tasks may wish to consult Appendix A of \citet{velivckovic2022clrs} for an overview of key ways to interact with CLRS---these will apply for CLRS-Text as well. The only change necessary to make in our default generator script for CLRS-Text is to indicate which part(s) of the trace should be printed---and in which format---for the newly added algorithm.

\section{Training and evaluation}

\begin{figure*}
    \centering
    \includegraphics[width=\linewidth]{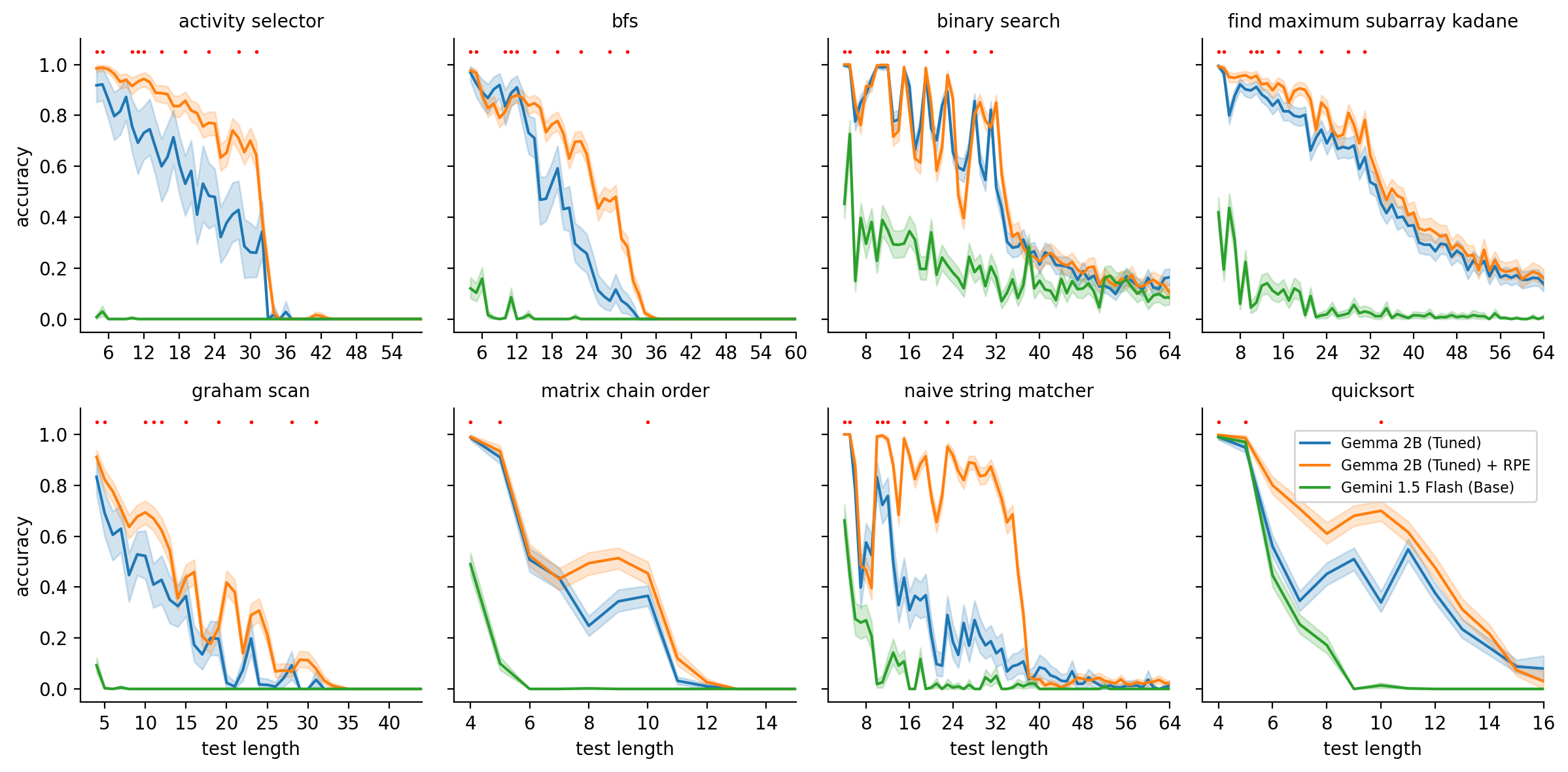}
    \caption{Resampling test results of variants of Gemma 2B, and Gemini 1.5 Flash, on various problem sizes. Gemma 2B variants were explicitly trained on CLRS-Text tasks---the training set sizes are denoted by red dots---and are evaluated zero-shot. Gemini 1.5 Flash is a pre-trained general-purpose model, evaluated in a two-shot manner. This plot only shows results on eight representative algorithms due to space constraints---the detailed plots for all thirty algorithms are available in Appendix \ref{app:full}.}
    \label{fig:results}
\end{figure*}

For the main part of our evaluation, we follow a rigorous out-of-distribution setup, wherein we pre-train a Gemma 2B model \citep{team2024gemma} using the standard next-token prediction objective on specific problem sizes across all thirty algorithms provided with CLRS by default. In the spirit of foundation models \citep{bommasani2021opportunities}, our pre-training follows the style of a \emph{generalist reasoner}---building a single multi-task model capable of executing all thirty algorithms from textual prompts simultaneously. 

To demonstrate how CLRS-Text can help assert established results, we also pre-train a variant of the Gemma 2B model which uses \emph{randomised} positional embeddings (RPE)---already shown by \citet{ruoss2023randomized} to yield better length generalisation for tasks along the Chomsky hierarchy.

This compares to relevant multi-task learning work on graph-based CLRS \citep{ibarz2022generalist,georgiev2024neural,bohde2024markov}, with a representational advantage in the language case: since the inputs and outputs are just textual tokens in CLRS-Text, exactly the same language model architecture can be used for all thirty tasks---whereas for the former papers, different encoder and decoder functions were necessary due to data and shape discrepancies.

Once trained, we evaluate our models zero-shot on randomly sampled CLRS-Text instances for each of the thirty algorithms, at every problem size that would fit in the model's context window, up to a maximum size of $n=64$. For each evaluation prompt, we extract the final array-like object that the language model produced in its output, and compare it to the ground-truth prompt via exact string match.

Note that resampling allows us to directly take into account the reasoning gap across multiple runs \citep{srivastava2024functional}, as we will never evaluate on static test data. Note also that this will evaluate both generalisation on in-distribution (training) problem sizes, interpolating out-of-distribution sizes and extrapolating out-of-distribution sizes.

For all our experiments, tool use (in the style of \citet{schick2024toolformer}), as well as the usage of code interpreters, is explicitly \emph{switched off}. This is due to the fact we would like to evaluate and improve \emph{base model} capabilities, without any confounding effects stemming from the tool's capabilities. 

These confounding effects could be particularly pronounced in the CLRS context, as the tasks from Introduction to Algorithms are standard hallmark of a computer science undergraduate degree---hence, their implementations are omnipresent on GitHub, and it is very likely that most frontier models have been trained on such source codes. We direct interested readers to \citet{mcleish2024benchmarking} for a detailed overview of the kinds of results achievable on a CLRS-Text variant when using a model with access to a code interpreter.

\section{Results and Discussion}

CLRS-Text lends itself naturally to many different forms of language model evaluation. We conduct one such evaluation, to provide an indication of how difficult it is to reliably incorporate algorithmic operations within language models.

We train two variants of the Gemma 2B model on all thirty tasks, using next-token prediction, at various training sizes. The first variant is identical to the original Gemma 2B, whereas the other employs randomised positional embeddings \cite{ruoss2023randomized} with $L=10,000$ (compared with Gemma's context length of $N=8,192$). 

The training sizes (see Appendix \ref{app:sizes}) are carefully chosen for each algorithm such that we can evaluate generalisation in both the interpolation and extrapolation regimes, taking into account Gemma 2B's context length. Note that only a single model is trained across all thirty tasks---mirroring the multi-task setup of \citet{ibarz2022generalist} for the domain of text. 

Evaluation is performed zero-shot, recording average accuracy across five resampled test sets of $125$ samples for every problem size fitting in context. Each example is assessed using exact string match on the final array produced.

While general-purpose frontier models are not necessarily out-of-distribution with respect to these problems or even these sizes, we believe it is useful to report the performance of such models on our test sets, to be able to internalise the gap these models have to cross before they will be capable to solve CLRS-Text tasks reliably. For this purpose, we include two-shot evaluation results for Gemini 1.5 Flash \citep{reid2024gemini}, which is a distilled, fast version of the frontier model Gemini 1.5 Pro.

We provide results for eight representative algorithms in Figure \ref{fig:results}; see Appendix \ref{app:full} for all thirty plots. While our results indicate that the use of randomised positional embeddings improves generalisation---as shown by \citet{ruoss2023randomized}---their effects still taper off quickly in the extrapolation regime. It is worth noting that, on the graph variant of CLRS, multi-task reasoners easily generalise to $4\times$ the input sizes seen at training time \citep{ibarz2022generalist}, whereas on CLRS-Text, language models barely extrapolate at all. We suspect that the autoregressive nature of LLMs is to blame---for reasoning problems, one can often predict multiple parts of the output at once rather than having to predict them one token at a time, which GNNs can easily exploit. We highlight this as an important future work direction.

In all cases, the pre-trained general-purpose model is not capable of achieving comparable results to the fine-tuned Gemma, indicating there exists a clear improvement in general-purpose algorithmic reasoning which is attainable with smaller models, that we may manage to make available to our frontier models in the future.



\bibliography{main}
\bibliographystyle{icml2024}

\newpage
\appendix
\onecolumn
\section{Algorithmic trace visualisations}
\label{app:samples}

\begin{figure*}[h]
    \includegraphics[width=\linewidth]{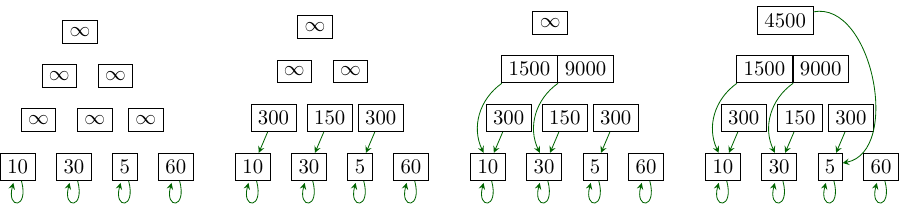}
    \textbf{\texttt{\textcolor{olivegreen}{matrix\_chain\_order:\\
    p: [10 30 5 60], initial\_trace: [[0 0 0 0], [0 0 0 0], [0 0 0 0], [0 0 0 0]]\\
    trace | s:}\\
    \textcolor{echonavy}{[[0 0 0 0], [0 0 1 0], [0 0 0 2], [0 0 0 0]] |} \textcolor{vividred}{[[0 0 0 0], [0 0 1 3], [0 0 0 2], [0 0 0 0]]}}}
    \caption{{\bf Top:} The \emph{graph} algorithmic trace of optimising the order of multiplications in a chain of matrices, for multiplying matrices of size $(10\times 30)(30\times 5)(5\times 60)$, assuming a $O(n^3)$-time multiplication algorithm (reprinted from \citet{velivckovic2022clrs}). {\bf Bottom:} The same algorithmic trace, represented \emph{textually}, by using our provided CLRS-Text generator. The model receives the input matrix sizes (\texttt{p}) and the initial value of the pointers (\texttt{initial\_trace}), using which it is prompted to predict the trace of gradually determining optimal orders of multiplying various subchains of the original chain of matrices. Note that, in our default data generator, we do not store intermediate numbers of operations---only the pointers are preserved in the trace.}
    \label{fig:mat}
\end{figure*}

\begin{figure*}[h]
    \includegraphics[width=\linewidth]{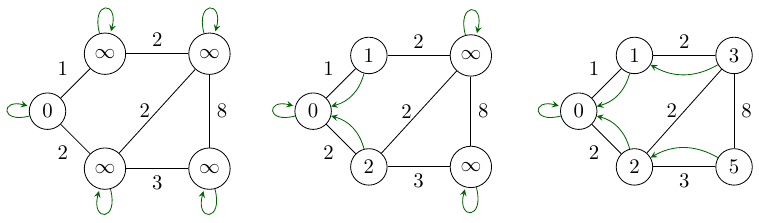}
    \textbf{\texttt{\textcolor{olivegreen}{bellman\_ford:\\
    s: 0, A: [[0 1 2 0 0], [1 0 0 2 0], [2 0 0 2 3], [0 2 2 0 8], [0 0 3 8 0]], initial\_trace:[0 1 2 3 4]\\
    trace | pi:}\\
    \textcolor{echonavy}{[0 0 0 3 4] |} \textcolor{vividred}{[0 0 0 1 2]}}}
    \caption{{\bf Top:} The \emph{graph} algorithmic trace of finding single-source shortest paths (from node zero) using the Bellman-Ford algorithm, for a given undirected weighted graph (reprinted from \citet{velivckovic2022clrs}). {\bf Bottom:} The same algorithmic trace, represented \emph{textually}, by using our provided CLRS-Text generator. The model receives the source node identity (\texttt{s}), the weighted adjacency matrix (\texttt{A}) and the initial value of the predecessor pointers (\texttt{initial\_trace}), using which it is prompted to predict the trace of gradually recomputing predecessor pointers until all single-source shortest paths are found. Note that, in our default data generator, we do not store intermediate path lengths---only the pointers are preserved in the trace.}
    \label{fig:graph}
\end{figure*}

\begin{figure*}[h]
    \includegraphics[width=\linewidth]{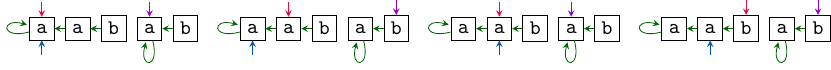}
    \textbf{\texttt{\textcolor{olivegreen}{naive\_string\_matcher:\\
    string: [0 0 0 1 1], key: [0 0 1 0 1], initial\_trace: 0\\
    trace | s:}\\
    \textcolor{echonavy}{0, 1 |} \textcolor{vividred}{1}}}
    \caption{{\bf Top:} The \emph{graph} algorithmic trace of finding the first occurence of the string \texttt{ab} inside the string \texttt{aab} (reprinted from \citet{velivckovic2022clrs}). {\bf Bottom:} The same algorithmic trace, represented \emph{textually}, by using our provided CLRS-Text generator. The model receives the identifier of which character belongs to which string (\texttt{string}), the value of each character (\texttt{key}) and the initial value of the position at which the match is queried (\texttt{initial\_trace}). Using this information, the model is prompted to predict the trace of gradually adjusting the querying position until a full match is found.}
    \label{fig:string}
\end{figure*}

Following the examples from \citet{velivckovic2022clrs}, we now provide additional trajectory examples captured by CLRS-Text, for a representative dynamic programming (Figure \ref{fig:mat}), graph (Figure \ref{fig:graph}) and string (Figure \ref{fig:string}) algorithm.

\section{Training sizes chosen for Gemma 2B}
\label{app:sizes}

In order to avoid overwhelming Gemma 2B's context window and to allow for a wide range of interpolation and extrapolation problem sizes at test time, we carefully choose the problem sizes for each algorithm used to generate the pre-training set for our Gemma 2B variants. Specifically, we leverage the sizes outlined in Table \ref{tab:train}.

The final pre-training set used by Gemma 2B in our experiments is constructed by taking $10,000$ randomly-sampled prompts of each training size.

\begin{table*}[h]
    \centering
    \begin{tabular}{ll}
    \toprule
    {\bf Algorithm} & {\bf Training sizes}\\
    \midrule
        Articulation points & $[4, 5, 10, 11, 12, 15, 19]$ \\
        Activity selector & $[4, 5, 10, 11, 12, 15, 19, 23, 28, 31]$\\
        Bellman-Ford & $[4, 5, 10, 11, 12, 15, 19, 23, 28, 31]$\\
        Binary search & $[4, 5, 10, 11, 12, 15, 19, 23, 28, 31]$\\
        Breadth-first search & $[4, 5, 10, 11, 12, 15, 19, 23, 28, 31]$\\
        Bridges & $[4, 5]$\\
        Bubble sort & $[4, 5, 10]$\\
        DAG shortest paths & $[4, 5, 10, 11, 12, 15, 19]$\\
        Depth-first search & $[4, 5, 10, 11, 12, 15, 19, 23]$\\
        Dijkstra & $[4, 5, 10, 11, 12, 15, 19, 23, 28]$\\
        Find Maximum Subarray & $[4, 5, 10, 11, 12, 15, 19, 23, 28, 31]$\\
        Floyd-Warshall & $[4, 5, 10]$\\
        Graham scan & $[4, 5, 10, 11, 12, 15, 19, 23, 28, 31]$\\
        Heapsort & $[4, 5, 10]$\\
        Insertion sort & $[4, 5, 10, 11, 12, 15, 19, 23, 28, 31]$\\
        Jarvis' march & $[4, 5, 10, 11, 12]$\\
        Kruskal's algorithm & $[4, 5, 10]$\\
        Knuth-Morris-Pratt & $[4, 5, 10, 11, 12, 15, 19, 23, 28, 31]$\\
        Longest common subsequence & $[4, 5, 10]$\\
        Matrix chain multiplication & $[4, 5, 10]$\\
        Minimum & $[4, 5, 10, 11, 12, 15, 19, 23, 28, 31]$\\
        Na\"{i}ve string matcher & $[4, 5, 10, 11, 12, 15, 19, 23, 28, 31]$\\
        Optimal binary search tree & $[4, 5, 10]$\\
        Prim's algorithm & $[4, 5, 10, 11, 12, 15, 19, 23, 28]$\\
        Quickselect & $[4, 5, 10, 11, 12, 15, 19, 23, 28, 31]$\\
        Quicksort & $[4, 5, 10]$\\
        Segments intersect & $[4, 5, 10, 11, 12, 15, 19, 23, 28, 31]$\\
        Strongly connected components & $[4, 5, 10, 11, 12, 15]$\\
        Task scheduling & $[4, 5, 10, 11, 12, 15, 19, 23, 28, 31]$\\
        Topological sort & $[4, 5, 10, 11, 12, 15, 19, 23]$\\
        \bottomrule
    \end{tabular}
    \caption{The training sizes employed for generating the pre-training set used for Gemma 2B models in our study. Note that, for Segments intersect, the size parameter does not influence the generated prompt.}
    \label{tab:train}
\end{table*}

\section{Results on all thirty algorithms}
\label{app:full}

\begin{figure}
    \centering
    \includegraphics[width=\linewidth]{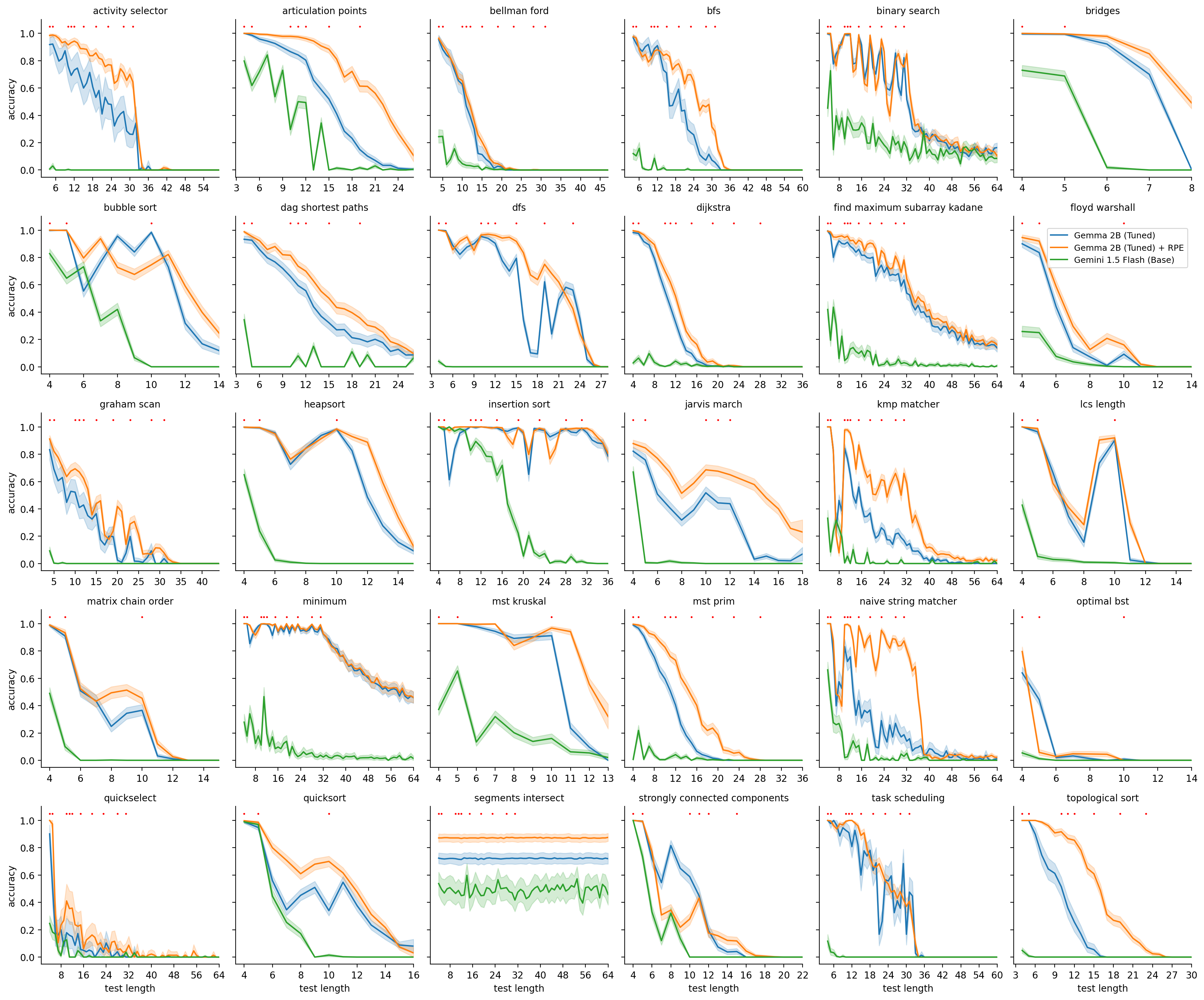}
    \caption{Resampling test results of variants of Gemma 2B, and Gemini 1.5 Flash, on various problem sizes. Gemma 2B variants were explicitly trained on CLRS-Text tasks---the training set sizes are denoted by red dots---and are evaluated zero-shot. Gemini 1.5 Flash is a pre-trained general-purpose model, evaluated in a two-shot manner.}
    \label{fig:full_res}
\end{figure}

For reasons of brevity, we were unable to show in the main paper the results of our models on all thirty algorithms presently released in CLRS-Text. Accordingly, these results---following identical conventions to Figure \ref{fig:results}---may be found in Figure \ref{fig:full_res}.

\end{document}